\documentclass[a4paper,11pt,twocolumn,twoside]{article}
\usepackage{sepln_en}
\usepackage{fullname}
\usepackage[utf8]{inputenc}
\usepackage{tabularx}
\usepackage{adjustbox}
\usepackage{subcaption}
\usepackage{booktabs}
\usepackage{url}
\usepackage{subcaption,booktabs,multirow,comment,graphicx,lipsum} 
\usepackage{comment}
\usepackage[para,online,flushleft]{threeparttable}
\usepackage{tikz}
\usepackage{listings}
\usepackage{soul}
\usepackage{framed}
\usepackage{enumitem}
\usepackage{algorithm}
\usetikzlibrary{arrows,shapes.geometric,arrows.meta,bending,chains,backgrounds,fit}

\lstnewenvironment{verbatim}{\lstset{breaklines,basicstyle=\ttfamily}}{}

\input epsf

\title{Language Independent Stance Detection: Social Interaction-based Embeddings and Large Language Models}

\author {\textbf{Joseba Fernandez de Landa,} \textbf{Rodrigo Agerri}\\
HiTZ Center - Ixa, University of the Basque Country UPV/EHU\\
joseba.fernandezdelanda@ehu.eus, rodrigo.agerri@ehu.eus\\}

\seplntranstitle{Detección de Stance Independiente del Idioma: Representaciones Vectoriales basadas en Interacciones Sociales y Grandes Modelos de Lenguaje}

\seplnclave{Detección de Stance, Multilingualismo, Redes Sociales, Interacciones, Modelos de Lenguaje, Procesamiento del Lenguaje Natural.}

\seplnresumen{La gran mayoría de los trabajos sobre la detección de \emph{stance} (posicionamiento) se han centrado en clasificación de texto, incluso cuando los datos se recolectan de redes sociales como Twitter. Este artículo aborda la tarea de detección de \emph{stance} haciendo énfasis, además de en los datos textuales de los mensajes, en los datos de interacción disponibles en las redes sociales. Proponemos un nuevo método para representar información social como \emph{amigos} y \emph{retuits} generando embeddings relacionales, es decir, representaciones vectoriales densas basadas en pares de interacción. Nuestros experimentos en siete conjuntos de datos públicamente disponibles y para cuatro idiomas (catalán, euskera, español e italiano) demuestran que la combinación de los embeddings relacionales con métodos textuales ayuda a mejorar el rendimiento, obteniendo resultados del estado del arte en seis de los siete escenarios de evaluación, superando otras aproximaciones basadas en grandes modelos de lenguaje u otros enfoques basados en interacciones como DeepWalk o node2vec.}

\seplnkey{Stance Detection, Multilinguality, Social Networks, Interactions, Large Language Models, Natural Language Processing.}

\seplnabstract{The large majority of the research performed on stance detection has been focused on developing more or less sophisticated text classification systems, even when many benchmarks are based on social network data such as Twitter. This paper aims to take on the stance detection task by placing the emphasis not so much on the text itself but on the interaction data available on social networks. More specifically, we propose a new method to leverage social information such as \emph{friends} and \emph{retweets} by generating Relational Embeddings, namely, dense vector representations of interaction pairs. Our experiments on seven publicly available datasets and four different languages (Basque, Catalan, Italian, and Spanish) show that combining our relational embeddings with discriminative textual methods helps to substantially improve performance, obtaining state-of-the-art results for six out of seven evaluation settings, outperforming strong baselines based on Large Language Models, or other popular interaction-based approaches such as DeepWalk or node2vec.}

\firstpageno{1}

\begin{document}


\setlength\titlebox{22cm} 

\label{firstpage} \maketitle

%

\section{Introduction}

Stance detection consists of identifying the viewpoint or attitude expressed by a piece of text with respect to a given target. With the enormous popularity  of social networks, users spontaneously share their opinions on social media, generating a valuable resource to investigate stance. This means that research on stance has a social impact, for example, to help addressing misinformation on vaccines, or to better understand public opinion about topics such as climate change or migration. Furthermore, stance detection is considered an important intermediate task for fact-checking \cite{Augenstein2021TowardsEF} or fake news detection\cite{pomerleau2017fake}.

The SemEval 2016 task on stance detection in Twitter \cite{mohammad-etal-2016-semeval} presented a dataset with tweets expressing FAVOR, AGAINST and NEUTRAL stances with respect to five different targets, a trend followed by many other researchers \cite{derczynski2017semeval,taule18,zotova2021semi,Hardalov2021FewShotCS}. However, despite many of them using Twitter-based source data, the large majority address the task by considering only the textual content of tweets \cite{augenstein-etal-2016-stance,Schiller2020StanceDB,hardalov-etal-2021-cross,li-etal-2021-improving-stance,ghosh2019stance,kuccuk2020stance,sobhani2017dataset,glandt2021stance}.

This shortcoming has been addressed by proposing new datasets \cite{cignarella2020sardistance,vaxxstance2021} that include different languages and social interaction data, such as \emph{retweets} or \emph{friends}. Although these new datasets have facilitated the development of new techniques for stance detection considering also interaction data, most of them employ manually engineered features tailored to each specific data type \cite{DeepReading,WordUp,QMUL-SDS}, making it difficult to generalize across languages and targets. Recently, significant attention has been directed towards the use of Large Language Models (LLMs) as few-shot learners \cite{brown2020}. However, the success of in-context learning techniques using LLMs has been mostly limited to English benchmarks such as SemEval 2016 \cite{taranukhin-etal-2024-stance,gatto-etal-2023-chain,zhangInvestigatingChainofthoughtChatGPT2023}, probably because the pre-training of the large majority of publicly available LLMs has been focused mostly on English.

This paper focuses on stance detection of tweets by placing emphasis on the interaction data commonly available in social media. More specifically, we propose a new method to leverage social information such as friends and retweets by generating Relational Embeddings, namely, dense vector representations of interaction pairs. The development of our new method allows us to make the following contributions to language independent stance detection: (i) a new method to represent and exploit interaction data, such as \emph{friends} and/or \emph{retweets}, by generating relational embeddings based on one-to-one relations; (ii) comprehensive experiments on seven publicly available datasets and four different languages different to English show that our relational embeddings behave robustly across different targets and languages without any specific manual engineering; (iii) combining our method with text-based discriminative classifiers helps to systematically improve their results, outperforming also ensembles of pre-trained language models \cite{Giorgioni2020UNITORS} or strong in-context learning baselines using Large Language Models \cite{taranukhin-etal-2024-stance}; (iv) we empirically demonstrate that our new Relational Embeddings clearly outperform popular graph-based approaches to encode interaction data, such as DeepWalk or node2vec; (v) exhaustive ablation and error analyses show that the method used to obtain the \emph{retweet} data and the size of the users community is crucial for state-of-the-art performance using our technique; (vi) the new generated datasets with interaction data and code are publicly available \footnote{\url{https://github.com/joseba-fdl/relational_embeddings/}}.

Finally, while this paper is focused on stance detection, we believe that our Relational Embeddings can be successfully applied to a large number of Computational Social Science and NLP tasks based on social media, especially those related to political ideology, misinformation, and hate speech, but also for health-related applications such as the detection of early signs of epidemic outbreaks \cite{Martin-Corral2022.11.15.22282355}.

\section{Related work}\label{sec:rel_work}

Recent studies have demonstrated that using LLMs on Stance Detection tasks can provide significant performance increases \cite{zhang2023automatic,zhangInvestigatingChainofthoughtChatGPT2023}. Furthermore, combining the application of LLMs with Chain-of-Thought (CoT) prompting \cite{weiChainThoughtPrompting2022}, and in-context learning in which the model generates intermediate reasoning steps to arrive at a final prediction, has also helped to substantially improve results \cite{kojima2022large,wang2023selfconsistency,gatto-etal-2023-chain}.

Despite their high capabilities, the application of LLMs still faces several challenges, such as dealing with cases of implicit stance or avoiding hallucinations, even when employing advanced prompting strategies such as CoT reasoning \cite{gatto-etal-2023-chain}. To address these limitations, Stance Reasoner \cite{taranukhin-etal-2024-stance} improves the CoT method by including examples and reasoning as background knowledge to achieve generalizable predictions across different targets. However, these approaches are only focused on English.

Additionally, most stance detection research and datasets released do not include interaction data, despite being collected from social media sources such as Twitter. \namecite{kuccuk2020stance} lists stance-annotated datasets for 11 languages, whereas recent work on cross-domain and cross-lingual stance provide experimentation for 16 datasets and 15 languages \cite{hardalov-etal-2021-cross,Hardalov2021FewShotCS}. The focus, however, remains on the textual content of the tweets. This trend has recently changed with the release of, to the best of our knowledge, two datasets which, in addition to the stance labeled tweets, include interaction data such as \emph{retweets} and \emph{friends}: SardiStance \cite{cignarella2020sardistance} and VaxxStance \cite{vaxxstance2021}.

The winner \cite{DeepReading} of the SardiStance shared task \cite{cignarella2020sardistance} used a weighted voting ensemble that combined two inputs: (a) psychological, sentiment and \emph{friends} distances as features used to learn an XGBoost \cite{friedman2001greedy} model, with (b) text classifiers based on the Transformer architecture \cite{Devlin19}. Other systems combined textual data (emoticons, special characters, and word embeddings) with 2 dimensions extracted from the interactions distance matrix using Multidimensional Scaling (MDS) \cite{TextWiller}, or friendship-based graphs created with DeepWalk \cite{deepwalk} and various types of textual embeddings \cite{QMUL-SDS}.

The VaxxStance shared task \cite{vaxxstance2021} provided textual and interaction data (\emph{friends} and \emph{retweets}) to study stance detection on vaccines in Basque and Spanish. The one system that systematically outperformed the baselines \cite{WordUp} manually engineered a large number of features based on stylistic, tweet, and user data, lexicons, dependency parsing, and network information, which were specifically developed for these datasets and languages.

The most recent approaches tackling unsupervised stance detection using social media interactions as features use the force-directed algorithm \cite{Fruchterman1991GraphDB} or UMAP \cite{McInnes2018UMAPUM}. These algorithms transform interaction frequency vectors into features, reducing huge interaction matrices into low-dimensional features. \namecite{darwish2020unsupervised} use both the force-directed algorithm and UMAP for unsupervised stance detection of Twitter users. UMAP is also used to get interaction-based features for automatically tagging Twitter users' stance \cite{Stefanov2020PredictingTT} and to explore political polarization in Turkey \cite{Rashed2021EmbeddingsBasedCF}. 

Other works are based on node2vec \cite{Grover2016node2vecSF} for user profiling and extracting user features for abuse detection \cite{mishra-etal-2018-author} and also for sentiment, stance and hate speech detection \cite{del-tredici-etal-2019-shall}. Commonly used algorithms for building interaction-based models like DeepWalk \cite{deepwalk} and node2vec are based on generating Random Walks. However, those randomly generated walks create artificial interactions that may not occur in the gathered interaction pairs. Furthermore, selecting the structure of the random walks and deciding the number of context users to be predicted needs to be manually modeled and adapted.

In contrast to previous work based on in-context learning with LLMs, supervised text classification or interaction-based methods such as DeepWalk or node2vec, our Relation Embeddings method provides dense interaction-based representations of users, focusing on real interaction pairs. The training process is designed to predict a target user receiving a \emph{retweet} or a \emph{follow} from a source user, each instance an item-to-item prediction instead of context-to-item (CBOW) or item-to-context (Skip-gram) prediction. Additionally, we focus on all the interaction pairs, without generating artificial random interactions to train the model or manually selecting the most salient users.

\section{Method}\label{sec:method}

We proposed a new method to generate vector-based representations of interactions in social networks, such as \emph{friends} and \emph{retweets}. These new representations, which we refer to as \emph{Relational Embeddings} (RE), are then leveraged to propose two methods to perform stance detection: (i) building classifiers using just our relational embeddings (\S \ref{sec:relational-classifier}) and, (ii) combining RE with various classifiers based on textual data (\S \ref{sec:comb-meth}).

\subsection{Relational Embeddings}\label{sec:relational_embeddings}

In this paper, the type of interactions used is \emph{retweets} and \emph{friends}, which are seen as relations between two users, one generating the action (source) and the other receiving it (target). Thus, the actions of \emph{retweeting} or \emph{following} other users are considered interaction pairs. Generally, these interactions should help to reveal users' preferences by capturing meaningful information from their performative actions.

The first step in our method consists of gathering the interactions from the users included in the labeled data, namely, the one-to-one \emph{retweet} and \emph{follow} actions between the users/authors of the tweets. It should be noted that a set of \emph{retweet} and \emph{follow} interactions can consist of independent one-to-one actions without direct relation between them. This is why in our model we consider each interaction pair as a single instance without any preprocessing or modification.

Using this interaction data, our model is then trained in an unsupervised manner to predict, in each instance, a target user from a given source user. Note that the instances used as input are real interaction pairs, namely, they do not correspond to sparse interaction frequency matrices or neighbors arising from interaction networks or without generating artificial ones as random walks do.

\begin{figure}[!ht]
	\centering
\includegraphics[scale=0.3]{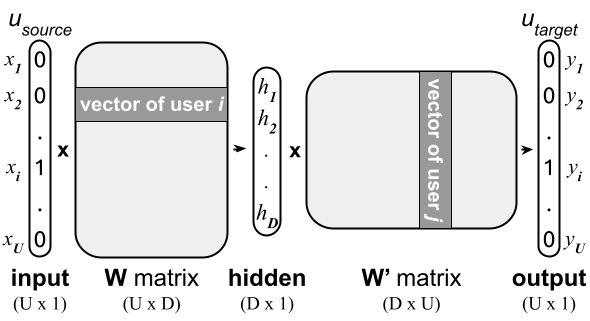} 
\caption{\footnotesize One hidden layer artificial neural network.}
\label{fig:arch}
\end{figure}

In order to obtain our relational representations, we use a single hidden-layer neural network (Figure \ref{fig:arch}). The network is used to train a dense interaction representation model using the \emph{friends} and/or \emph{retweet} based data. Each user is encoded as a one-hot vector of size $U$, where $U$ is the number of users among interaction pairs ($I$) in a specific dataset. Given a one-hot vector $U$, the aim of the single hidden-layer feedforward neural network consists of predicting the target user. The dimensions of the hidden layer ($D$) determine the size of relational vectors representing the target user, which correspond to the number of learned features. During training, the weights $W$ and $W'$ are modified to minimize the loss function due to backpropagation. According to Equation \ref{eq:u2v_2}, the summation goes over all the interaction pairs ($I$) in the training corpus, computing the log probability of correctly predicting the target user ($u_{target}$) from the source user ($u_{source}$) for each interaction ($i$). The training process is done by sub-sampling the most frequent instances and with negative sampling \cite{mikolov2013efficient}. Finally, the $W$ matrix is used to retrieve the interaction vectors representing each user, generating the relational embedding, from which the relation vector for each user is obtained. In this model, users with similar interactions should have similar representations, turning many interaction pairs into dense relational representations of $D$ dimensions.

\begin{equation} \small
\frac{1}{I}\sum_{i=1}^{I} \log p(u_{target}|u_{source})
\label{eq:u2v_2}
\end{equation}

\subsection{Interaction-based Classifier with Relational Embeddings} \label{sec:relational-classifier}

Our first system consists of a linear classifier taking as input only the relational embeddings described in the previous section. Building such a system will allow us to understand the performance of the generated Relational Embedding models on their own.

\begin{figure}[!ht]
	\centering
\includegraphics[scale=0.5]{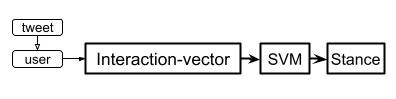}
\caption{\footnotesize Relational Embeddings + SVM model architecture.}\label{fig:relemb-svm}
\end{figure}

Each of the tweets from a dataset will be represented by its author's (user) relational vector, which represents the interactions of its author. By doing so, we effectively project the relations of the author to the tweet level, generating a link between the relational data and the stance labels. In this step, some users may be repeated across the data, but their assigned stance label will be that of the corresponding tweet. It should be noted that, although possible, it is quite uncommon to have a user with tweets labeled differently across the data. Thus, each tweet is converted into relational vectors, represented by the specific user's vector weights in the relational embedding model. Those users not present in the model are represented as vectors of zeros. This is usually due to the inability to retrieve user interaction data, either because the user has disappeared from Twitter or because their profiles are kept private. As shown in Figure \ref{fig:relemb-svm}, the final relational vectors for each tweet are used to train an SVM (RBF kernel) classifier without any additional input. 

\subsection{Combining Textual and Interaction Data}\label{sec:comb-meth}

In order to combine textual and interaction data, we use both the texts conveyed by a given user and its associated social media interactions as data input. More specifically, we propose an ensemble method combining textual and interaction-based features before learning. This ensemble method is applied to combine interaction and textual representations with SVM and Transformers. 

\begin{figure}[!ht]
	\centering
\includegraphics[scale=0.5]{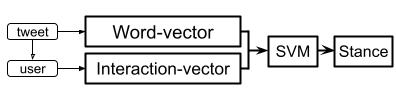}
\caption{\footnotesize SVM-based combined models architecture.}\label{fig:comb-svm}
\end{figure}

In Figure \ref{fig:comb-svm}, we obtain a FastText (\textit{FTEmb}) dense or TFIDF sparse word vectors to represent each tweet. The interaction embedding of each author will then be concatenated to the textual vector, adding a vector of zeros to the textual vector if no relational information is available. Finally, the concatenation of the textual and relational vectors is used to learn an SVM (RBF kernel) model.

\begin{figure}[!ht]
	\centering
\includegraphics[scale=0.5]{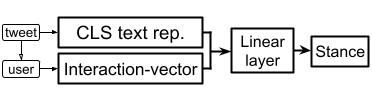}
\caption{\footnotesize Transformer-based combined model architecture.}\label{fig:comb-trans}
\end{figure}

As shown in Figure \ref{fig:comb-trans}, Transformer models and interaction embeddings are combined by concatenating user vectors from the interaction embeddings with the Transformer's CLS representations of the tweets. When there is no user information related to interactions, a vector of zeros is concatenated to the CLS representation. Finally, we add a linear classification layer on top of the CLS token vector concatenated with the relational vector and fine-tune the model in an end-to-end manner.

\section{Stance Detection Datasets}\label{sec:datasets}

In order to be able to experiment with the interaction data, the datasets should include, in addition to the labeled textual data, interactions of the users that published the tweets, such as each user's \emph{friends} and \emph{retweets}. To the best of our knowledge, there are currently only two publicly available datasets with such contextual information, namely, SardiStance \cite{cignarella2020sardistance} and VaxxStance \cite{vaxxstance2021}. 

We would like to note that we tried to obtain user information for SemEval 2016 \cite{mohammad-etal-2016-semeval} and other English datasets \cite{lai2020brexit,Conforti2020WillTheyWontTheyAV,Glandt2021StanceDI}, without much success. In the case of SemEval 2016, perhaps the most popular stance detection dataset, we only managed to retrieve less than 30\% of users, not nearly enough for meaningful experimentation with interaction data. For other datasets, it was simply not possible to extract the tweets based on the IDs published. However, we did manage to extract the interactions (over 80\%) for the Catalonia Independence Corpus \cite{zotova2021semi} (CIC).

The number of labeled tweets and their distribution between train and test sets can be seen in Table \ref{tab:data_analysis}. This choice of data offers a varied relation user-tweets (very low in SardiStance, quite high in CIC), which would also allow us to test the robustness of our method.

\begin{table}[h!]
\centering
\resizebox{7.5cm}{!}{
\begin{tabular}{@{}lrrrrrr@{}} 
\toprule
& \multicolumn{3}{c}{Tweets} & \multicolumn{3}{c}{Interactions} \\ \cmidrule{2-7} \cmidrule{2-7}
& Train & Test & Total & Users & RTs & Friend \\ \midrule
SE2016 & 2,914 & 1,249 & 4,163 & - & - & - \\
C-ca & 8,038 & 2,010 &10,048 & 691 & \multirow{4}{*}{10M\textsuperscript{†}} & \multirow{4}{*}{24M\textsuperscript{†}} \\
C-ca*& 8,056 & 1,992 &10,048 & 691 & & \\
C-es & 8,036 & 2,011 &10,047 & 334 & & \\
C-es* & 8,016 & 2,031 &10,047 & 334 & & \\ \midrule
S & 2,132 & 1,110 &3,242 & 2,827 & 575K & 3M \\\midrule
V-eu & 1,072 & 312 &1,384 & 210 & 190K& 170K \\
\multirow{2}{*}{V-es} & \multirow{2}{*}{2,003} & \multirow{2}{*}{694} & \multirow{2}{*}{2,697} & \multirow{2}{*}{1,675} & 9K & \multirow{2}{*}{2.1M} \\ 
& & & & & 552K\textsuperscript{†} & \\ \bottomrule
\end{tabular}
}
\caption{\footnotesize Datasets: SE2016 (SemEval 2016), C (CIC), S (SardiStance), V (VaxxStance); * means no overlap of users across train and test; RT (retweets). {†} mark represents supplementary interaction-based data added by us.}
\label{tab:data_analysis}
\end{table}

\subsection{SemEval 2016}\label{sec:semeval2016}
The SemEval2016 dataset \cite{mohammad-etal-2016-semeval}, consists of English tweets labeled for
stance (AGAINST, FAVOR, and NONE) from an initial pool of 2M collected tweets. 
In the supervised track, 4163 tweets are provided with manual annotation for five targets: ``Atheism'', ``Climate Change is a Real Concern'', ``Feminist Movement'', ``Hillary Clinton'', and ``Legalization of Abortion''. The annotated tweets were ordered by their timestamps. Interestingly, in addition to stance, annotations are provided to express whether the target is explicitly mentioned in the tweet. Finally, no user-based information is included in the datasets, just the tweets.

\subsection{SardiStance}
This dataset contains tweets in Italian about the Sardines movement \cite{cignarella2020sardistance}. In addition to the textual data, this dataset also provides social and user information, such as the authors' friends and the retweets. We were unable to extract any supplementary data because both tweet and user identifiers are encrypted.

\subsection{VaxxStance}
The dataset was independently collected for two languages: Basque and Spanish. No user overlap across train and test sets occurs in the data. Interactions, such as \emph{friends} from the users and \emph{retweets} made to the labeled tweets, are also included. The Basque version (VaxxStance-eu) also includes retweets retrieved from the users' timelines, as there are few tweets retrieved from the labeled tweets. In order to get more interaction-based data, we extracted 552K supplementary retweets from the users' timelines of the Spanish subset (VaxxStance-es), emulating the extra retweet collection as the authors did for the Basque version.

\subsection{Catalonia Independence Corpus}
The Catalonia Independence Corpus includes coetaneous tweets in both Spanish and Catalan \cite{zotova-etal-2020-multilingual}, is multilingual, quite large (10K tweets), and reasonably balanced. In the original CIC data, 92.50\% of the users in the Catalan set occur also in the test set, whereas for Spanish the proportion is even higher, namely, 99.72\%. In order to avoid any possible overfit to the author's style, a second version of the dataset \cite{zotova2021semi} distributes the tweets in such a way that their authors do not appear across the training, development, and test sets (CIC*).

\section{Baselines}\label{sec:baselines}

We compare state-of-the-art in-context learning with LLMs methods with supervised text-based statistical and Transformer encoder-only classifiers. Furthermore, as explained in Section \ref{sec:method}, supervised textual methods are combined with interaction-based features extracted from the Twitter social network. Thus, in addition to our Relational Embeddings, we also experiment with interaction-based methods such as node2vec and DeepWalk for direct comparison with our proposal.

\paragraph{In-context learning with LLMs}\label{sec:llm-meth} 

While traditional supervised text-based methods fine-tune a language model on some training data, in-context learning is based on proving the language model with optimized prompts \cite{brown2020}. The benefits are two-fold: no training is required, just a handful of examples (few-shot), and by guiding the language model to learn in-context it facilitates the generalization capability of the model to unseen targets. Taking into account their multilinguality claims, we prove the LLMs Mistral 7B v0.2 instruct \cite{jiang2023mistral} and Llama3 8B instruct \cite{Dubey2024TheL3} with two different types of prompts plus examples and their translations into the specific language of the dataset (full examples provided in Annex \ref{sec:appendix:prompts}).

\noindent - \textbf{Few-Shot} A prompt including 6 in-context examples, two for each label, on targets different to the ones included in the evaluation set. The examples are carefully crafted to include features that would maximize the generalization of the models \cite{taranukhin-etal-2024-stance}.

\noindent - \textbf{Stance Reasoner} The few-shot prompt is augmented with chain-of-thought \cite{weiChainThoughtPrompting2022} to generate intermediate reasoning steps that lead to a label prediction \cite{taranukhin-etal-2024-stance}.

\paragraph{Supervised text-based methods}\label{sec:txt-meth}

In order to combine interaction-based with textual-only supervised approaches, we choose three commonly used text classification methods for stance detection \cite{aldayel2020stance,kuccuk2020stance,hardalov-etal-2021-cross,zotova2021semi} as baselines to compare with our relational embeddings models: 

\noindent - \textbf{Word Embeddings}: We use FastText CommonCrawl models trained using the C-BOW architecture and 300 dimensions on a vocabulary of 2M words \cite{grave2018learning}. For classification, each tweet is represented as the average of its word vectors \cite{kenter-etal-2016-siamese} which is then used to train an SVM (RBF kernel) classifier.

\noindent - \textbf{TFIDF}: TFIDF (Term Frequency Inverse Document Frequency) vectorization is applied in order to reduce word vector dimensionality by lowering the impact of words that occur too frequently in the selected corpus. TFIDF vectorizing is applied over the text of the tweets, selecting most salient features and reducing sparsity. The obtained TFIDF vectors are used to learn a SVM (RBF kernel) model.

\noindent - \textbf{XLM-RoBERTa}: We use XLM-RoBERTa \cite{conneau2019unsupervised} for text classification, a masked encoder-only language model pre-trained for 100 languages on 2.5 TB of CommonCrawl text. This model has been widely tested for multilingual stance detection with state-of-the-art performance \cite{aldayel2020stance,ghosh2019stance,kuccuk2020stance,DeepReading,zotova2021semi}.

\begin{table*}[!ht]
  \centering
  \footnotesize
  \begin{threeparttable}
    \begin{tabularx}{\textwidth}{lXXXXXr}
      \toprule
      \textbf{Model} & LA & AT & CC & FM & HC & \textbf{Overall} \\
      \midrule
      \multicolumn{7}{c}{\textit{Supervised}} \\
      \hspace{0.5cm}{TFIDF}      & 62.2 & 53.8 & 41.6 & 55.5 & 58.1 & 62.9 \\
      \hspace{0.5cm}{FTEmb}      & 60.4 & 59.8 & 40.0 & 53.8 & 47.7 & 65.8 \\
      \hspace{0.5cm}{XLM-R}      & \underline{63.7} & \underline{\textbf{72.4}} & \underline{45.5} & \underline{58.3} & \underline{73.9} & \underline{72.5} \\
      \midrule
      
      \multicolumn{7}{c}{\textit{Few-shot with LLMs}} \\
      
      \textbf{Few-Shot} \\
      \hspace{0.5cm}{Mistral 7B}     & 66.7 & 48.9 & \underline{65.4} & 64.7 & 76.1 & 70.7  \\
      \hspace{0.5cm}{LLaMA3 8B}      & 64.3 & 56.1 & 57.9 & 68.6 & \underline{\textbf{78.3}} & 70.8  \\
      
      \textbf{Stance Reasoner} \\
      \hspace{0.5cm}{Mistral 7B}     & \underline{\textbf{72.3}} & \underline{72.0} & 60.0 & \underline{\textbf{74.0}} & 76.1 & \underline{\textbf{75.4}} \\
      \hspace{0.5cm}{LLaMA3 8B}      & 63.2 & 69.3 & 64.9 & 68.0 & 76.8 & 73.7 \\
      
      \midrule 
      \hspace{0.5cm}{Previous SOTA}  &   \textbf{74.3} & 44.6 & \textbf{65.7} & 65.0 & 71.3 & 75.1 \\
      
      \bottomrule
    \end{tabularx}
  \end{threeparttable}
  \caption{
    \label{tab:results_semeval}
    Overall $F_1$ macro results on the SemEval 2016 task 6a dataset \cite{mohammad-etal-2016-semeval} and for each target in the test split, namely, 
    LA - \textit{Legalization of Abortion},
    AT - \textit{Atheism},
    CC - \textit{Climate Change is a Real Concern},
    FM - \textit{Feminist Movement},
    HC - \textit{Hillary Clinton}
    . The best results are highlighted in bold. The best results for each group in underlined. Few-shot with LLMs: our implementation with prompts from Taranukhin et al. \shortcite{taranukhin-etal-2024-stance}.
  }
\end{table*}

\paragraph{Interaction-based Methods}\label{sec:inter-meth}

In addition to the Relational Embeddings proposed in Section \ref{sec:relational_embeddings}, we also compare with two commonly used graph-based approaches to extract user representations of the tweets' authors which are then used to train an SVM (RBF kernel) classifier, as described in Section \ref{sec:relational-classifier}.

\noindent - \textbf{DeepWalk} \cite{deepwalk}: Given a node(s) in the network, this algorithm learns feature representations to predict their context or neighbors. In this item-to-context (Skip-gram) predicting task, the neighbors to predict may be artificially generated by simulating random walks among the connected nodes. 

\noindent - \textbf{Node2vec} \cite{Grover2016node2vecSF}: Similar to DeepWalk, it adds two parameters to control the structure of the network during the generated random walks. The control parameters focus on the probability of revisiting points and on the probability of visiting further points.

\section{Experiments}

The experimental setup consists of (i) evaluating the textual baselines on SemEval 2016 and, (ii) experimenting with interaction data on seven datasets including three different topics (Independence of Catalonia, anti-vaxxers, Sardines movement) and four languages (Basque, Catalan, Italian and Spanish). This choice of datasets will allow us to first evaluate, using SemEval 2016, popular supervised textual classifier methods based on statistical and Transformers encoder-only models with respect to state-of-the-art in-context learning techniques with Large Language Models (LLMs). Once we have established the performance of each textual classifier for English in SemEval 2016, we will then combine the textual supervised methods with interaction data (as explained in Section \ref{sec:comb-meth}) and compare them with in-context learning with LLMs in CIC, VaxxStance and SardiStance.

\subsection{Experimental Settings}

Regarding which interactions to choose to complement textual classifiers, for each dataset (CIC, SardiStance, and VaxxStance) three different types of interaction-based embeddings were trained, based on the data source: (i) \emph{retweets}, (ii) \emph{friends} and, (iii) \emph{mixed} embeddings.

\emph{Retweets} are used to share specific content from other users' publications. The reiteration of these actions may demonstrate attachment to a user or its content, actively showing the specific preferences of the source user. Furthermore, although retweet actions are more likely to encode latent information related to community or polarization \cite{conover2011political,Zubiaga2019PoliticalHI}, we also wanted to include \emph{friends} related data, which is a result of a \emph{following} action. This passive action allows the source user to be aware of what is being said without sharing or promoting any content. Finally, we also combine both \emph{retweets} and \emph{friends} in a \emph{mixed} representation to test whether merging passive and active interaction types in the same interaction space helps to embed social information.

The best interaction type to build the embeddings for each dataset was chosen by evaluating them with the classifiers built with interaction-based representations (RE, DW and N2V) via 5-fold cross-validation on the training data. The results showed that RE \emph{retweet} was the best interaction data for CIC and VaxxStance-eu, whereas the \emph{mixed} embeddings were the best for SardiStance and VaxxStance-es. With respect to DW and N2V, \emph{retweets} were best for CIC and VaxxStance, while \emph{mixed} performed better for SardiStance. 

The procedure to choose the rest of the hyperparameters (dimensionality of the interaction embeddings etc.) for every method is described in Appendix \ref{sec:appendix}. Finally, as it is customary for this task, despite training and predictions being done for the 3 classes, evaluation is performed by calculating the averaged F1-score over the AGAINST and FAVOR classes \cite{mohammad-etal-2016-semeval}.

\begin{table*}[!ht]
  \centering
  \footnotesize
    \begin{tabularx}{\textwidth}{lXXXXXXXr}
      \toprule
      \textbf{Model} & C-ca & C-ca* & C-es & C-es* & S & V-es & V-eu    & \textit{\textbf{avg.}}   \\
      \midrule

      \multicolumn{9}{c}{\textit{Few-shot with LLMs }}  \\
      \textbf{Few-shot} \\
      \hspace{0.5cm}{Mistral 7B}                    & 20.1 & 22.0 & 28.8 & 28.5 & 48.1 & 63.8 & 34.1 & 35.1 \\
      \hspace{0.5cm}{Mistral 7B (translated)}       & 28.8 & 32.9 & 32.4 & 33.5 & 56.3 & 64.0 & 32.7 & 40.1 \\
      \hspace{0.5cm}{LLaMA3 8B}                     & 49.9 & 56.9 & 48.8 & 51.2 & 59.0 & 50.6 & \underline{47.5} & 52.0 \\
      \hspace{0.5cm}{LLaMA3 8B (translated)}        & \underline{55.4} & \underline{59.7} & \underline{49.5} & \underline{52.0} & \underline{63.1} & 54.1 & 43.3 & \underline{53.9} \\
      \textbf{Stance Reasoner} \\
      \hspace{0.5cm}{Mistral 7B}                    & 47.4 & 46.6 & 44.2 & 45.8 & 50.9 & \underline{65.0} & 41.6 & 48.8 \\
      \hspace{0.5cm}{Mistral 7B (translated)}       & 49.0 & 51.1 & 43.0 & 44.5 & 55.4 & 64.0 & 41.2 & 49.7 \\
      \hspace{0.5cm}{LLaMA3 8B}                     & \underline{50.6} & \underline{55.7} & \underline{51.6} & \underline{52.7} & 60.4 & 59.1 & \underline{46.8} & \underline{53.9} \\
      \hspace{0.5cm}{LLaMA3 8B (translated)}        & 49.6 & 53.2 & 48.9 & 50.7 & \underline{60.5} & 59.9 & 44.0 & 52.4 \\
      \midrule
      
      \multicolumn{9}{c}{\textit{Supervised}}  \\
      \textbf{Text-based}  \\
      \hspace{0.5cm}{TFIDF}         & 75.3 & 71.6 & 73.7 & 73.1 & \underline{63.4} & 76.5 & \underline{54.4} & \underline{69.7} \\
      \hspace{0.5cm}{FTEmb}         & 61.6 & 62.6 & 57.4 & 59.8 & 56.1 & 71.3 & 47.7 & 59.5 \\
      \hspace{0.5cm}{XLM-R}         & \underline{77.6} & \underline{74.6} & \underline{74.2} & \underline{73.9} & 57.2 & \underline{82.5} & 41.2 & 68.7 \\
      \textbf{Interaction-based}  \\
      \hspace{0.5cm}{RE (ours)}     & \underline{\textbf{82.2}} & \underline{70.2} & \underline{85.2} & \underline{84.4} & \underline{71.7} & \underline{85.5} & \underline{48.4} & \underline{75.4} \\
      \hspace{0.5cm}{N2V}           & 69.7 & 61.8 & 65.9 & 49.3 & 65.7 & 76.2 & 29.0 & 59.7 \\ 
      \hspace{0.5cm}{DW}            & 69.1 & 68.0 & 66.5 & 64.0 & 66.4 & 79.5 & 25.7 & 62.7 \\ 

      \textbf{Combined}  \\
      \hspace{0.5cm}{RE + FTEmb}    & \underline{\textbf{82.2}} & 69.2 & 88.6 & \underline{\textbf{87.7}} & 74.0 & 89.1 & 73.2 & 80.6 \\
      \hspace{0.5cm}{RE + TFIDF}    & \underline{\textbf{82.2}} & \underline{\textbf{80.2}} & \underline{\textbf{92.5}} & 86.6 & \underline{\textbf{74.6}} & \underline{\textbf{90.2}} & \underline{75.3} & \underline{\textbf{83.1}} \\
      \hspace{0.5cm}{RE + XLM-R}    & 78.8 & 75.9 & 76.8 & 77.2 & 60.2 & 81.5 & 51.8 & 71.7 \\ 
      \hspace{0.5cm}{N2V + FTEmb}   & 71.7 & 63.7 & 71.3 & 57.7 & 70.3 & 80.4 & 48.4 & 66.2 \\ 
      \hspace{0.5cm}{N2V + TFIDF}   & 77.6 & 74.1 & 80.1 & 72.6 & 70.9 & 85.8 & 54.1 & 73.6 \\ 
      \hspace{0.5cm}{N2V + XLM-R}   & 77.2 & 73.7 & 72.9 & 74.6 & 56.3 & 80.6 & 39.2 & 67.8 \\ 
      \hspace{0.5cm}{DW + FTEmb}    & 72.9 & 67.3 & 71.6 & 72.9 & 67.7 & 82.7 & 48.5 & 69.1 \\ 
      \hspace{0.5cm}{DW + TFIDF }   & 79.2 & 75.9 & 81.1 & 79.7 & 70.1 & 86.0 & 54.7 & 75.2 \\ 
      \hspace{0.5cm}{DW + XLM-R}    & 78.0 & 75.1 & 73.5 & 72.8 & 55.6 & 78.7 & 47.3 & 68.7 \\

      \midrule
      Previous SOTA & 74.7 & 74.9 & 74.7 & 71.8 & 74.4 & 89.1 & \textbf{77.7} & 76.7 \\
      \bottomrule
    \end{tabularx}
  \caption{
    \label{tab:results_multylang}
    $F_1$ macro scores \cite{mohammad-etal-2016-semeval}. Best results highlighted in bold; best results per group underlined. Few-shot with LLMs: our implementation using prompts from Taranukhin et al. \shortcite{taranukhin-etal-2024-stance} with Mistral v0.2 7B and Llama 3 8B, both instruct. Supervised: using interaction-based systems (RE, N2V, and DW), text-based systems (FTEmb, TF-IDF and XLM-RoBERTa) and their combinations. Previous SOTA: CIC \cite{zotova-etal-2020-multilingual,zotova2021semi}, SardiStance \cite{DeepReading} and VaxxStance \cite{WordUp}.
  }
\end{table*}

\subsection{Evaluation Results}

The evaluation of stance detection systems on the SemEval 2016 dataset (Table \ref{tab:results_semeval}) highlights the effectiveness of few-shot approaches leveraging large language models (LLMs). In particular, the Stance Reasoner configuration with Mistral and LlaMA3 consistently surpasses the fine-tuned XLM-RoBERTa's performance, outperforming also the previous state-of-the-art (SOTA) results obtained by Stance Reasoner \cite{taranukhin-etal-2024-stance}. These results demonstrate the strength of in-context learning for English using modern LLMs.

Moving on to the experiments with datasets with available interaction data, in Table \ref{tab:results_multylang} we report the results of evaluating stance detection systems on CIC (Catalan and Spanish) SardiStance (Italian) and VaxxStance (Basque and Spanish). If we look first at the text-only approaches we observe that, unlike for SemEval 2016, supervised approaches clearly outperform few-shot in-context learning with LLMs methods. We hypothesize that this is due to the English-centric development of LLMs. The `translated' rows refer to the results obtained when the tweets included in the few-shot configuration are translated to the target languages. These results show that while LLMs benefit from having the examples included in the prompt in the target languages, the gains are much lower with the Stance Reasoner prompt, in which the CoT reasoning component plays a crucial role.

The results using interaction-based approaches underline the usefulness of encoding social interactions for stance detection. This is particularly true of the Relational Embeddings, which outperform every text-based method using only relational vectors and SVM as a learning algorithm. Relational Embeddings also outperform interaction-based alternatives like node2vec and DeepWalk, emphasizing the value of modeling interactions as relational pairs rather than arbitrary random walks. 

Most importantly, combining text-based supervised methods with our Relational Embeddings systematically helps to improve results for every language and dataset, obtaining in most cases state-of-the-art results. Thus, for SardiStance and VaxxStance, Relational Embeddings achieve performance comparable to SOTA without manual tuning or extensive additional resources, in contrast to previous systems that relied heavily on manually engineered features and external data \cite{WordUp}. Finally, 
the combination of Relational Embeddings with TFIDF (RE+TFIDF) consistently shows the best performance, even outperforming systems leveraging pre-trained language models and state-of-the-art LLMs. These findings demonstrate the potential of combining supervised textual and relational features for robust and language-independent stance detection.

\section{Discussion}\label{sec:discussion}

\begin{figure}[!htb]
  \centering
  \begin{subfigure}[b]{0.49\linewidth}
    \includegraphics[width=\linewidth]{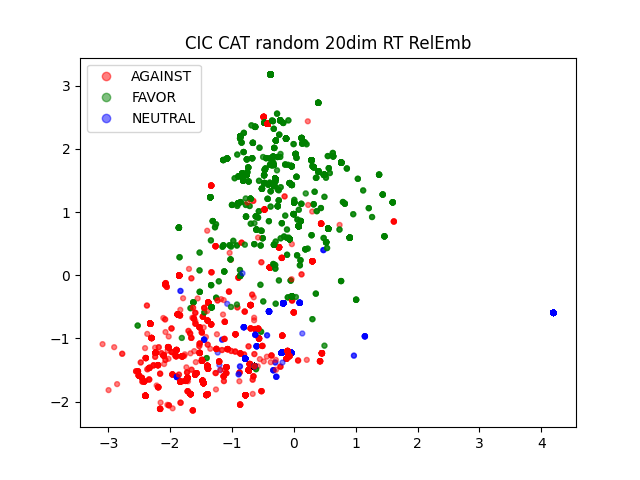}
    \caption{CIC-ca*}\label{fig:emb_rep_CIca-r}
  \end{subfigure}
  \begin{subfigure}[b]{0.49\linewidth}
    \includegraphics[width=\linewidth]{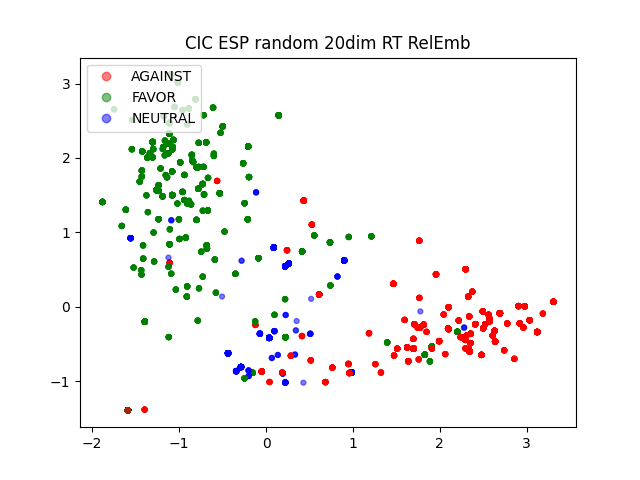}
    \caption{CIC-es*}\label{fig:emb_rep_CIes-r}
  \end{subfigure}
  \begin{subfigure}[b]{0.49\linewidth}
    \includegraphics[width=\linewidth]{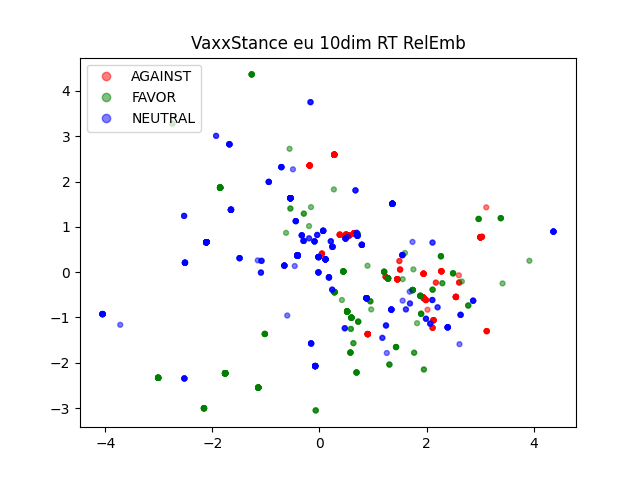}
    \caption{VaxxStance-eu}\label{fig:emb_rep_VaXeu}
  \end{subfigure}
  \begin{subfigure}[b]{0.49\linewidth}
    \includegraphics[width=\linewidth]{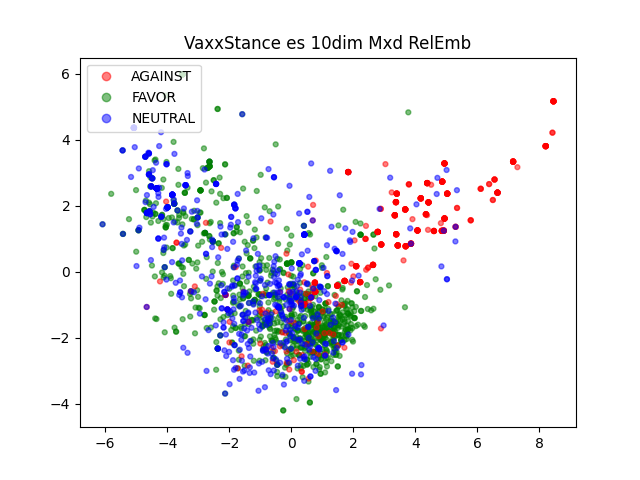}
    \caption{VaxxStance-es}\label{fig:emb_rep_VaXes}
  \end{subfigure}
  \begin{subfigure}[b]{0.49\linewidth}
    \includegraphics[width=\linewidth]{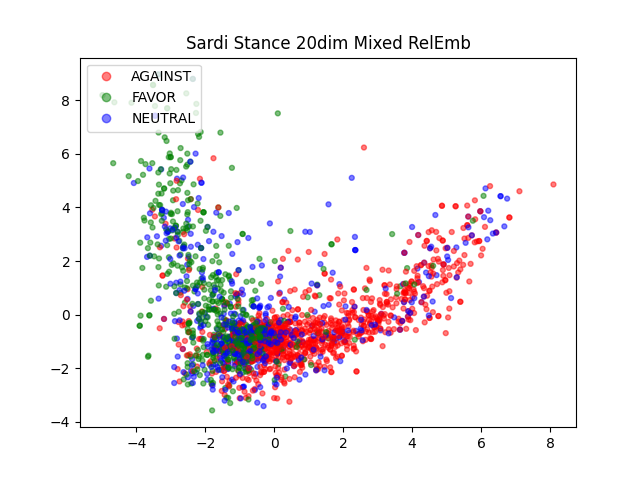}
    \caption{SardiStance}\label{fig:emb_rep_Sardi}
  \end{subfigure}
  \caption{\footnotesize Relational embedding representation of training set users (PCA dimension reduction to 2).}
  \label{fig:emb_rep}
\end{figure}

In order to have a clearer understanding of the results obtained by the different interaction-based methods, we visualize their generated representations via PCA reduction. The reported results have shown that Relational Embeddings, which use actual interaction pairs rather than random walk methods like N2V and DW, lead to superior results. Visualizations of user representations, reduced to 2D with PCA, reveal a clear link between the readability of relational embeddings and system performance (Figure \ref{fig:emb_rep}).

CIC data shows distinct FAVOR and AGAINST communities, while SardiStance data displays more overlap. VaxxStance-eu embeddings struggle to separate stances, likely due to the small, highly interactive Basque Twitter community. Topics with strong political homophily, such as Catalan independence, generate clearer embeddings than less polarized or smaller datasets like VaxxStance.

Comparisons in Appendix \ref{sec:appendix} show that DW and N2V embeddings produce less coherent visualizations, reflecting their lower performance.

\section{Conclusion and Future Work}

Stance Detection is a challenging task as the selection of the target significantly influences the assignment of labels. For instance, the same piece of text may express support for one target while simultaneously opposing another, depending on the context and framing. This dependency makes it critical to perform the task in a generalizable way, as even a small change may affect the outcome. LLMs with in-context learning offer a promising approach to achieving generalizable Stance Detection. However, they face notable limitations, including a substantial error rate (approximately 25\%) and significant difficulties in handling non-English languages.

In this paper, we propose Relational Embeddings, a new method to represent interaction data such as \emph{retweets} and \emph{friends}. Relational Embeddings help to reduce the sparsity of interaction data by behaving like dense graphs, being able to embed information related to stance from different data sources without any manual engineering. While this technique is language-independent and fast to train and apply, our results demonstrate that Relational Embeddings behave robustly across different datasets, targets, and languages, substantially and consistently improving results by combining them with text-based supervised methods. The results show that using Relational Embeddings also outperforms most text classification baselines. Furthermore, a direct comparison with previous interaction-based approaches such as DeepWalk and node2vec shows the superiority of our approach.

The results and analysis performed show that we should pay more attention to social network data, aiming to address the shortcomings discussed by further researching different strategies to leverage such interaction data. Future work may include analyzing the Relational Embeddings performance on zero-shot and cross-lingual settings, moving on towards a method that, by using user-based Relational Embeddings, helps to drastically reduce the need of annotated data at tweet level.

\section*{Acknowledgements}

We are thankful to the following MCIN/AEI/10.13039/501100011033 projects: (i) DeepKnowledge (PID2021-127777OB-C21) and by FEDER, EU; (ii) Disargue (TED2021-130810B-C21) and European Union NextGenerationEU/PRTR; (iii) DeepMinor (CNS2023-144375) and European Union NextGenerationEU/PRTR.

\bibliographystyle{fullname}
\bibliography{anthology,custom}

\appendix
\section{Appendix}\label{sec:appendix}

The dimensionality for the interaction-based embeddings was chosen by training a SVM classifier via grid search with 5-fold cross-validation. Based on results reported in previous work on reducing huge interaction matrices into low dimensional features for stance detection \cite{darwish2020unsupervised,Stefanov2020PredictingTT}, dimensions for interaction-based embeddings were selected between 10 or 20 dimensions. The best performing RE for CIC and SardiStance were of dimension 20, while for VaxxStance they were of dimension 10. With respect to DW and N2V, the best dimensionality for CIC-es, CIC-es* and SardiStance was 10, whereas the best one for CIC-ca, CIC-ca* and VaxxStance correspond to 20. Moreover, for DW and N2V we set the usual default values for the hyperparameters for these algorithms: walks\_per\_node = 10, walk\_length = 80, window or context\_size = 10, and the optimization is run for a single epoch \cite{deepwalk,Grover2016node2vecSF}. Specifically for node2vec, we have set p=1 and q=0.5 in order to enhance network community-related information \cite{Grover2016node2vecSF}. 

Grid search with 5-fold cross-validation was also used to optimize C and Gamma hyperparameters for every SVM system (RE, DW, N2V, FTEmb, TFIDF, and all the combinations).

For XLM-RoBERTa, hyperparameter tuning was done by splitting the training set into training and development sets (80/20). Results on the development set allowed us to obtain the following hyperparameters: 128 maximum sequence length, 16 batch size, 2e-5 learning rate, and 5 epochs.

\begin{figure*}[ht]
  \centering
  \begin{subfigure}[b]{0.32\linewidth}
    \includegraphics[width=\linewidth]{fig/cic_cat_ran.png}
    \caption{CIC-ca* RE}\label{fig:emb_rep_CIca-r-bis}
  \end{subfigure}
  \begin{subfigure}[b]{0.32\linewidth}
    \includegraphics[width=\linewidth]{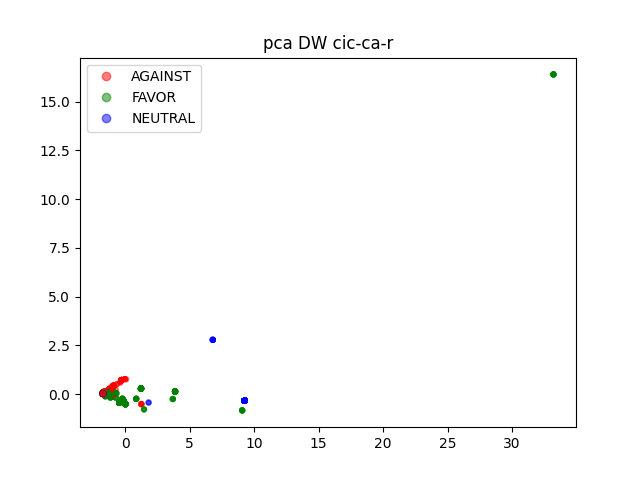}
    \caption{CIC-ca* DW}\label{fig:emb_rep_CIca-r_dw}
  \end{subfigure}
  \begin{subfigure}[b]{0.32\linewidth}
    \includegraphics[width=\linewidth]{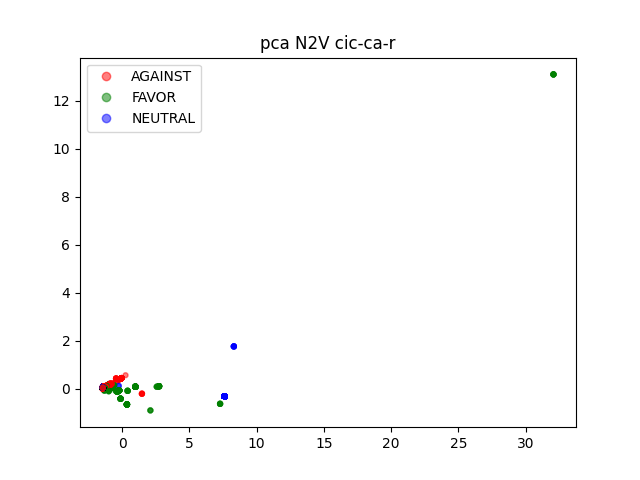}
    \caption{CIC-ca* N2V}\label{fig:emb_rep_CIca-r_n2v}
  \end{subfigure}
  
  \begin{subfigure}[b]{0.32\linewidth}
    \includegraphics[width=\linewidth]{fig/cic_esp_ran.png}
    \caption{CIC-es* RE}\label{fig:emb_rep_CIes-r-bis}
  \end{subfigure}
  \begin{subfigure}[b]{0.32\linewidth}
    \includegraphics[width=\linewidth]{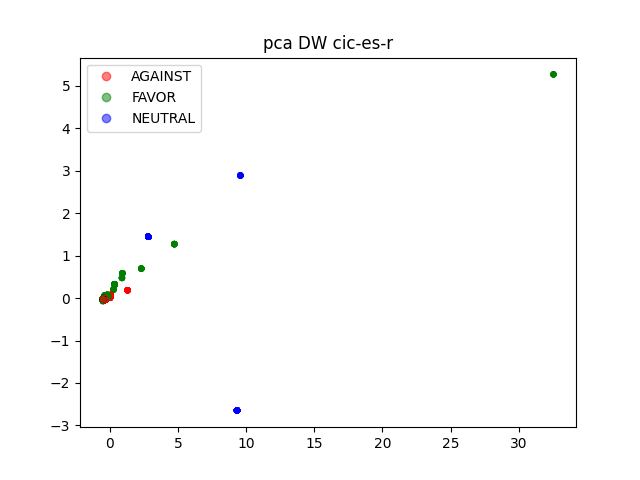}
    \caption{CIC-es* DW}\label{fig:emb_rep_CIes-r_dw}
  \end{subfigure}
  \begin{subfigure}[b]{0.32\linewidth}
    \includegraphics[width=\linewidth]{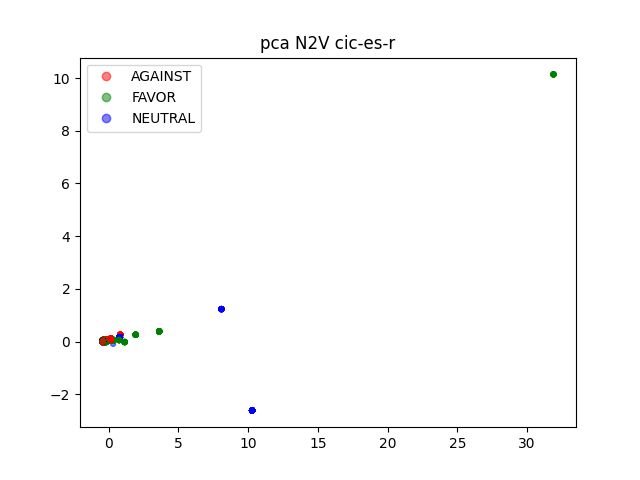}
    \caption{CIC-es* N2V}\label{fig:emb_rep_CIes-r_n2v}
  \end{subfigure}

  \begin{subfigure}[b]{0.32\linewidth}
    \includegraphics[width=\linewidth]{fig/vs_eu.png}
	\caption{VaxxStance-eu RE}\label{fig:emb_rep_VaXeu-bis}
  \end{subfigure}
  \begin{subfigure}[b]{0.32\linewidth}
    \includegraphics[width=\linewidth]{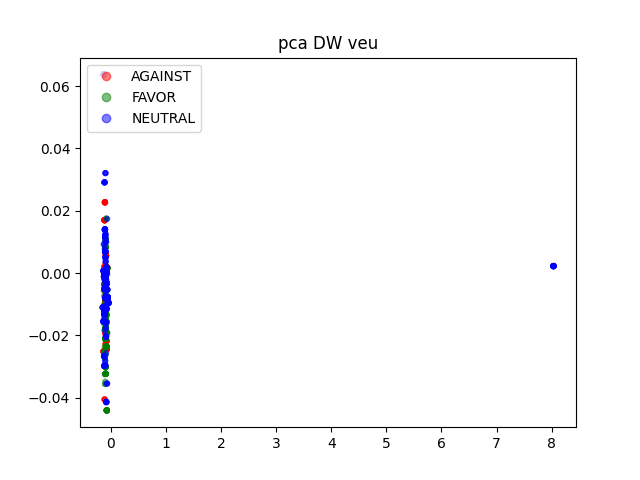}
    \caption{VaxxStance-eu DW}\label{fig:emb_rep_VaXeu_dw}
  \end{subfigure}
  \begin{subfigure}[b]{0.32\linewidth}
    \includegraphics[width=\linewidth]{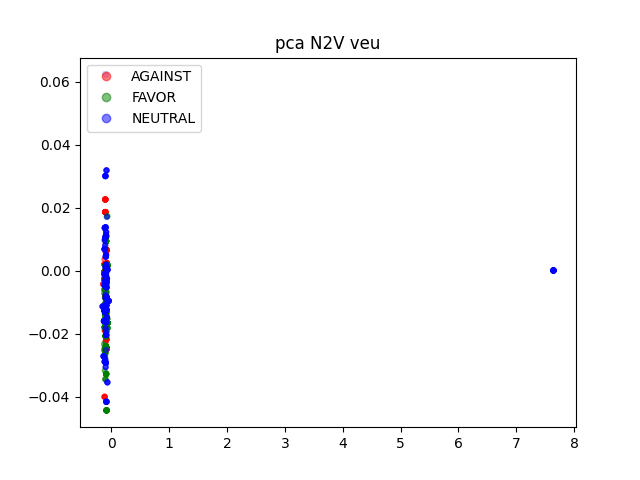}
    \caption{VaxxStance-eu N2V}\label{fig:emb_rep_VaXeu_n2v}
  \end{subfigure}
  
  \begin{subfigure}[b]{0.32\linewidth}
    \includegraphics[width=\linewidth]{fig/vs_es.png}
	\caption{VaxxStance-es RE}\label{fig:emb_rep_VaXes-bis}
  \end{subfigure}
  \begin{subfigure}[b]{0.32\linewidth}
    \includegraphics[width=\linewidth]{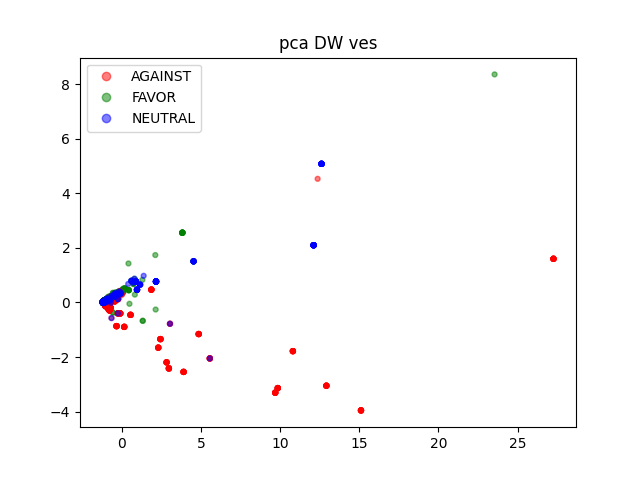}
    \caption{VaxxStance-es DW}\label{fig:emb_rep_VaXes_dw}
  \end{subfigure}
  \begin{subfigure}[b]{0.32\linewidth}
    \includegraphics[width=\linewidth]{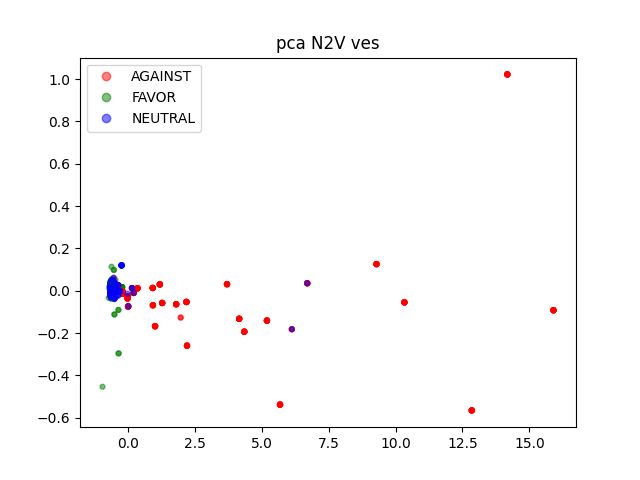}
    \caption{VaxxStance-es N2V}\label{fig:emb_rep_VaXes_n2v}
  \end{subfigure}

  \begin{subfigure}[b]{0.32\linewidth}
    \includegraphics[width=\linewidth]{fig/ss.png}
	\caption{SardiStance RE}\label{fig:emb_rep_Sardi-bis}
  \end{subfigure}
  \begin{subfigure}[b]{0.32\linewidth}
    \includegraphics[width=\linewidth]{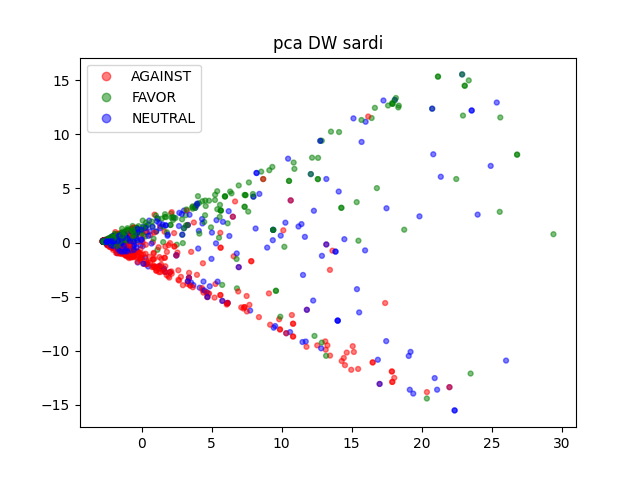}
    \caption{SardiStance DW}\label{fig:emb_rep_Sardi_dw}
  \end{subfigure}
  \begin{subfigure}[b]{0.32\linewidth}
    \includegraphics[width=\linewidth]{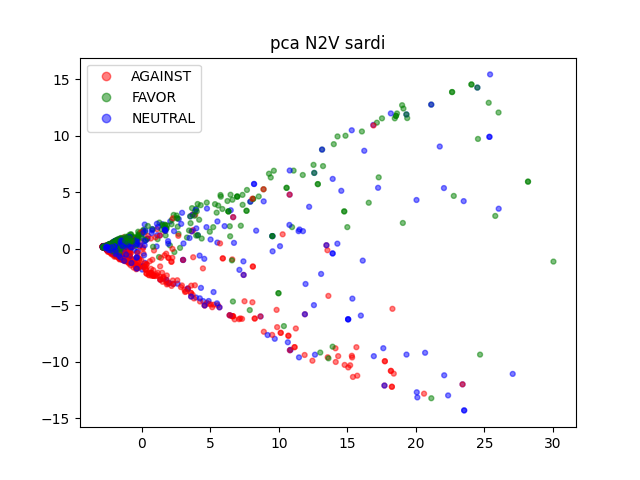}
    \caption{SardiStance N2V}\label{fig:emb_rep_Sardi_n2v}
  \end{subfigure}
  
  \caption{\footnotesize Comparison between Relational Embedding (RE), Deep Walk (DW) and Node2Vec (N2V) representations of training set users (PCA dimension reduction to 2).}
  \label{fig:re_and_dw}
\end{figure*}

Figure \ref{fig:re_and_dw} provides a direct comparison between the RE, DW and N2V user representations for each of the training datasets.

\newpage
\onecolumn
\section{Prompts}
\label{sec:appendix:prompts}

\subsection{Few-Shot Prompt}
\label{sec:appendix:prompts:few-shot}

Few-shot prompt with 6 examples developed by \namecite{taranukhin-etal-2024-stance}.
\begin{framed}
  \ttfamily \small \noindent
    Question: Consider the tweet in a conversation about the target, what could the tweet's point of view be towards the target?\newline
  The options are: \newline
  - against \newline
  - favor \newline
  - neutral \newline
  \newline
  tweet: <I'm sick of celebrities who think being a well-known actor makes them an authority on anything else. \#robertredford \#UN> \newline
  target: Liberal Values \newline
  stance: against \newline
  \newline
  tweet: <I believe in a world where people are free to move and choose where they want to live>\newline
  target: Immigration\newline
  stance: favor\newline
  \newline
  tweet: <I love the way the sun sets every day. \#Nature \#Beauty>\newline
  target: Taxes\newline
  stance: neutral\newline
  \newline
  tweet: <If a woman chooses to pursue a career instead of staying at home, is she any less of a mother?>\newline
  target: Conservative Party\newline
  stance: against\newline
  \newline
  tweet: <We need to make sure that mentally unstable people can't become killers \#protect \#US>\newline
  target: Gun Control\newline 
  stance: favor\newline
  \newline
  tweet: <There is no shortcut to success, there's only hard work and dedication \#Success \#SuccessMantra>\newline
  target: Open Borders\newline
  stance: neutral
\end{framed}

\newpage
\subsection{Stance Reasoner Prompt}
\label{sec:appendix:prompts:sr}

The Stance Reasoner prompt containing 6 examples and reasoning chains \cite{taranukhin-etal-2024-stance}.

\begin{framed}
  \ttfamily \small \noindent
    Question: Consider the tweet in a conversation about the target, what could the tweet's point of view be towards the target?\newline
  The options are: \newline
  - against \newline
  - favor \newline
  - neutral \newline
  \newline
  tweet: <I'm sick of celebrities who think being a well-known actor makes them an authority on anything else. \#robertredford \#UN> \newline
  target: Liberal Values \newline
  reasoning: the author is implying that celebrities should not be seen as authorities on political issues, which is often associated with liberal values such as Robert Redford who is a climate change activist -> the author is against liberal values \newline
  stance: against \newline
  \newline
  tweet: <I believe in a world where people are free to move and choose where they want to live>\newline
  target: Immigration\newline
  reasoning: the author is expressing a belief in a world with more freedom of movement -> the author is in favor of immigration.\newline
  stance: favor\newline
  \newline
  tweet: <I love the way the sun sets every day. \#Nature \#Beauty>\newline
  target: Taxes\newline
  reasoning: the author is in favor of nature and beauty -> the author is neutral towards taxes\newline
  stance: neutral\newline
  \newline
  tweet: <If a woman chooses to pursue a career instead of staying at home, is she any less of a mother?>\newline
  target: Conservative Party\newline
  reasoning: the author is questioning traditional gender roles, which are often supported by the conservative party -> the author is against the conservative party\newline
  stance: against\newline
  \newline
  tweet: <We need to make sure that mentally unstable people can't become killers \#protect \#US>\newline
  target: Gun Control\newline 
  reasoning: the author is advocating for measures to prevent mentally unstable people from accessing guns -> the author is in favor of gun control.\newline
  stance: favor\newline
  \newline
  tweet: <There is no shortcut to success, there's only hard work and dedication \#Success \#SuccessMantra>\newline
  target: Open Borders\newline
  reasoning: the author is in favor of hard work and dedication -> the author is neutral towards open borders\newline
  stance: neutral
\end{framed}

\newpage
\subsection{Translated tweets for Few-Shot and Stance Reasoner Prompts}
\label{sec:appendix:prompts:trans}

We retain the original prompts while translating the language of the 'tweet' into the specific target language.

\subsubsection{Basque}

\begin{framed}
  \ttfamily \small \noindent
    tweet: <Nazkatuta nago aktore ezagunak izateak beste edozein gaitan aditu bihurtzen dituela uste duten famatuekin. \#robertredford \#UN> \newline
    tweet: <Jendea aske mugitzeko eta bizi nahi duen lekua aukeratzeko munduan sinesten dut> \newline
    tweet: <Eguzkia egunero nola sartzen den maite dut. \#Natura \#Edertasuna> \newline
    tweet: <Emakume batek karrera bat jarraitzea aukeratzen badu etxean geratu beharrean, horrek ama txarragoa bihurtzen al du?> \newline
    tweet: <Buruko gaixotasunak dituzten pertsonak hiltzaile bihurtu ez daitezen ziurtatu behar dugu \#babestu \#AEB> \newline
    tweet: <Ez dago arrakastara daraman lasterbiderik, lana eta dedikazioa baino ez daude \#Arrakasta \#ArrakastarenGakoa> \newline
\end{framed}

\subsubsection{Catalan}
\begin{framed}
  \ttfamily \small \noindent
tweet: <Estic fart dels famosos que es pensen que ser un actor conegut els converteix en una autoritat en qualsevol altra cosa. \#robertredford \#UN> \newline
tweet: <Crec en un món on la gent sigui lliure de moure's i escollir on vol viure> \newline
tweet: <M'encanta com es pon el sol cada dia. \#Natura \#Bellesa> \newline
tweet: <Si una dona tria seguir una carrera professional en lloc de quedar-se a casa, és menys mare per això?> \newline
tweet: <Hem d'assegurar-nos que les persones mentalment inestables no puguin convertir-se en assassins \#protegir \#EUA> \newline
tweet: <No hi ha dreceres cap a l'èxit, només treball dur i dedicació \#Èxit \#MantraDelÈxit> \newline
\end{framed}

\subsubsection{Italian}
\begin{framed}
  \ttfamily \small \noindent
tweet: <Sono stufo dei personaggi famosi che pensano che essere un attore conosciuto li renda esperti di qualsiasi altro argomento. \#robertredford \#UN> \newline
tweet: <Credo in un mondo dove le persone siano libere di muoversi e scegliere dove vogliono vivere> \newline
tweet: <Amo il modo in cui il sole tramonta ogni giorno. \#Natura \#Bellezza> \newline
tweet: <Se una donna sceglie di perseguire una carriera invece di rimanere a casa, è meno madre per questo?> \newline
tweet: <Dobbiamo assicurarci che le persone mentalmente instabili non possano diventare assassini \#proteggere \#USA> \newline
tweet: <Non esistono scorciatoie per il successo, ci sono solo duro lavoro e dedizione \#Successo \#MantraDelSuccesso> \newline
\end{framed}

\subsubsection{Spanish}
\begin{framed}
  \ttfamily \small \noindent
tweet: <Estoy harto de los famosos que creen que ser un actor conocido los convierte en autoridad sobre cualquier otro tema. \#robertredford \#UN> \newline
tweet: <Creo en un mundo donde las personas sean libres de moverse y elegir dónde quieren vivir> \newline
tweet: <Me encanta cómo se pone el sol cada día. \#Naturaleza \#Belleza> \newline
tweet: <Si una mujer elige seguir una carrera en lugar de quedarse en casa, ¿es menos madre por ello?> \newline
tweet: <Necesitamos asegurarnos de que las personas mentalmente inestables no puedan convertirse en asesinos \#proteger \#EEUU> \newline
tweet: <No hay atajos hacia el éxito, solo trabajo duro y dedicación \#Éxito \#MantraDelÉxito> \newline
\end{framed}

\end{document}